\documentclass[11pt,a4paper]{article}
\usepackage{times,latexsym}
\usepackage{url}
\usepackage[T1]{fontenc}
\usepackage{booktabs}
\usepackage{appendix}
\usepackage{mathtools}
\usepackage{tabularx}
\usepackage[normalem]{ulem}
\usepackage{paralist}
\usepackage{comment}
\usepackage[normalem]{ulem}
\usepackage{subcaption}

\usepackage[acceptedWithA]{tacl2018v2}

\usepackage{xspace,mfirstuc,tabulary}

\newif\iftaclinstructions
\taclinstructionsfalse %
\iftaclinstructions
\renewcommand{\confidential}{}
\renewcommand{\anonsubtext}{(No author info supplied here, for consistency with
TACL-submission anonymization requirements)}
\newcommand{\instr}
\fi

\iftaclpubformat %

\else

\fi

\newcommand{\model}{SIFT\xspace}
\newcommand{\modell}{SIFT-Light\xspace}

\newcommand{\interalia}[1]{\citep[\emph{inter alia}]{#1}}

\usepackage{amsmath}
\usepackage{graphicx}
\usepackage{multirow}
\usepackage{hhline}
\usepackage{textcomp}

\definecolor{orange}{rgb}{1,0.5,0}
\definecolor{mdgreen}{rgb}{0.05,0.6,0.05}
\definecolor{mdblue}{rgb}{0,0,0.7}
\definecolor{dkblue}{rgb}{0,0,0.5}
\definecolor{dkgray}{rgb}{0.3,0.3,0.3}
\definecolor{slate}{rgb}{0.25,0.25,0.4}
\definecolor{gray}{rgb}{0.5,0.5,0.5}
\definecolor{ltgray}{rgb}{0.7,0.7,0.7}
\definecolor{purple}{rgb}{0.7,0,1.0}
\definecolor{lavender}{rgb}{0.65,0.55,1.0}

\newcommand{\com}[1]{}

\newcommand{\revcolor}{black}
\newcommand{\rev}[1]{\textcolor{\revcolor}{#1}}

\def\ve{{\mathbf{e}}}

\def\vh{{\mathbf{h}}}

\def\mW{{\mathbf{W}}}

\def\gN{{\mathcal{N}}}

\def\gR{{\mathcal{R}}}

\DeclarePairedDelimiter\abs{\lvert}{\rvert}%

\title{Infusing Finetuning with Semantic Dependencies}

\author{Zhaofeng Wu$^\spadesuit$ \quad
    Hao Peng$^\spadesuit$ \quad
    Noah A. Smith$^{\spadesuit\diamondsuit}$ \\
    $^\spadesuit$Paul G. Allen School of Computer Science \& Engineering, University of Washington \\
    $^\diamondsuit$Allen Institute for Artificial Intelligence \\
    \texttt{\{zfw7,hapeng,nasmith\}@cs.washington.edu}

}

\date{}

\begin{document}
\maketitle
\begin{abstract}
For natural language processing systems, two kinds of evidence support the use of text representations from neural language models ``pretrained'' on large unannotated corpora:  performance on application-inspired benchmarks \interalia{peters2018deep}, and the emergence of syntactic abstractions in those representations \interalia{tenney2019you}.  On the other hand, the lack of grounded supervision calls into question how well these representations can ever capture meaning \citep{bender-koller-2020-climbing}.  We apply novel probes to recent language models---specifically focusing on predicate-argument structure as operationalized by semantic dependencies \citep{ivanova2012did}---and find that, unlike syntax, semantics is not brought to the surface by today's pretrained models. %
We then use convolutional graph encoders to explicitly incorporate semantic parses into task-specific finetuning, yielding benefits to natural language understanding \rev{(NLU)} tasks in the GLUE benchmark.  This approach demonstrates the potential for general-purpose (rather than task-specific) linguistic supervision, above and beyond  conventional pretraining and finetuning.
Several diagnostics help to localize the benefits of our approach.\footnote{We release our code, pretrained models, and parsed semantic graphs for GLUE in four semantic formalisms at \url{https://github.com/ZhaofengWu/SIFT}.}
\end{abstract}

\begin{figure*}[th!]
	\centering
	\includegraphics[width=0.95\textwidth]{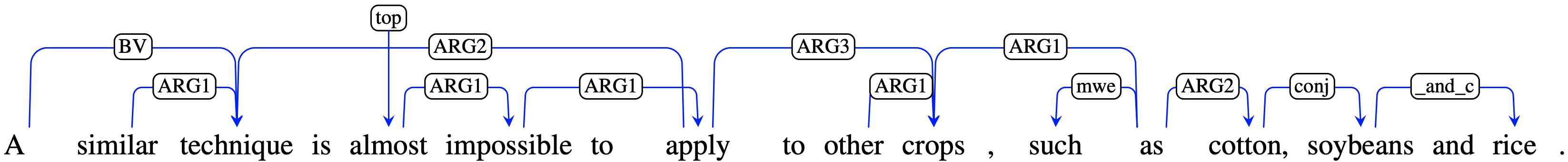}
	\includegraphics[width=0.95\textwidth]{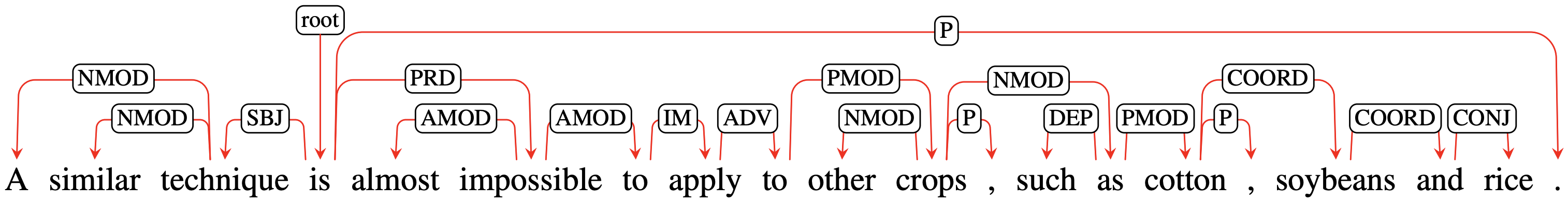}
	\caption{An example sentence in the DM (top, blue) and Stanford Dependencies (bottom, red) format, taken from \citet{oepen2015semeval} and \citet{ivanova2012did}.}
	\label{fig:dm-example}
\end{figure*}

\section{Introduction}
The past decade has seen a paradigm shift in how NLP systems are built, summarized as follows:
\begin{compactitem}
    \item Before, general-purpose linguistic modules (e.g., part-of-speech taggers, word-sense disambiguators, and many kinds of parsers) were constructed using supervised learning from linguistic datasets.  These were often applied as preprocessing to text as part of larger systems for information extraction, question answering, and other applications.
    \item Today, general-purpose representation learning is carried out on large, unannotated corpora---effectively a kind of unsupervised learning known as ``pretraining''---and then the representations are ``finetuned'' on application-specific datasets using conventional end-to-end neural network methods.
\end{compactitem}
The newer paradigm encourages an emphasis on corpus curation, scaling up pretraining, and translation of end-user applications into trainable ``tasks,'' purporting to automate most of the labor requiring experts (linguistic theory construction, annotation of data, and computational model design).  Apart from  performance improvements on virtually every task explored in the NLP literature, a body of evidence from probing studies has shown that pretraining brings linguistic abstractions to the surface, without explicit supervision \interalia{liu2019linguistic,tenney2019you,hewitt2019structural,goldberg2019assessing}.

There are, however, reasons to pause.  
First, some have argued from first principles that 
\rev{learning mappings from form to meaning is hard from forms alone~\citep{bender-koller-2020-climbing}.}\footnote{
\rev{
In fact, \citet{bender-koller-2020-climbing} argued that this is impossible for grounded semantics.
Our probing analysis, along with recent efforts~\citep{kovaleva-etal-2019-revealing,liu2019linguistic}, 
suggests that modern pretrained models are poor at surfacing predicate-argument semantics.
}}
Second, probing studies have focused more heavily on \emph{syntax}
than on \emph{semantics} (i.e., mapping of forms to abstractions of meaning intended  by people speaking in the world).  
\citet{tenney2019you} noted that the BERT model \citep{devlin2018bert} offered more to syntactic tasks like constituent and dependency relation labeling than semantic ones like Winograd coreference and semantic proto-role labeling.  \citet{liu2019linguistic} showed that pretraining did not provide much useful information for entity labeling or coreference resolution. \citet{kovaleva-etal-2019-revealing} found minimal evidence that the BERT attention heads capture FrameNet \citep{framenet} relations. We extend these findings in \S\ref{sec:probing}, showing that representations from the RoBERTa model \citep{liu2019roberta} are relatively poor at surfacing information for a predicate-argument semantic parsing probe, compared to what can be learned with finetuning, or what RoBERTa offers for \emph{syntactic} parsing.  The same pattern holds for BERT.  

Based on that finding, we hypothesize that semantic supervision may still be useful to tasks targeting natural language ``understanding.''  In \S\ref{sec:method}, we introduce \textbf{semantics-infused finetuning} (SIFT), inspired by pre-neural pipelines.  Input sentences are first passed through a semantic dependency parser.  Though the method can accommodate any graph over tokens, our implementation uses the DELPH-IN MRS-derived dependencies, known as ``DM'' \cite{ivanova2012did}, illustrated in Figure~\ref{fig:dm-example}.  The task architecture learned during finetuning combines the pretrained model (here, RoBERTa) with a relational graph convolutional  network (RGCN;~\citealp{schlichtkrull2018modeling}) that reads the graph parse.
Though the same graph parser can be applied at inference time (achieving our best experimental results), benefits to task performance are in evidence 
in a ``light'' model variant without inference time parsing and with the same inference cost as a RoBERTa-only baseline.

We experiment with the GLUE benchmarks (\S\ref{sec:experiments}), which target many aspects of natural language understanding \citep{wang2018glue}. 
\rev{Our model consistently improves over both base and large-sized RoBERTa baselines.}\footnote{\rev{RoBERTa-base and RoBERTa-large use the same pretraining data and only differ in the number of parameters.}} 
Our focus is not on achieving a new state of the art, but we note that SIFT can be applied orthogonally alongside other methods that have improved over similar baselines,
such as \citet{raffel2019exploring} and \citet{clark2020electra}, which used alternative pretraining objectives, and \citet{jiang2019smart}, which proposed an alternative finetuning optimization framework.
In \S\ref{sec:analysis},
we use the HANS \citep{mccoy2019right} and GLUE \citep{wang2018glue} diagnostics to better understand where our method helps on natural language inference tasks.  We find that our model's gains strengthen when finetuning data is reduced, and that our approach is more effective than alternatives that do not use the full labeled semantic dependency graph.

\section{Predicate-Argument Semantics as Dependencies} \label{sec:pas}

Though many formalisms and annotated datasets have been proposed to capture various facets of natural language semantics, here our focus is on predicates and arguments evoked by words in sentences.  Our experiments focus on the DELPH-IN dependencies formalism \citep{ivanova2012did}, commonly referred to as ``DM'' and derived from minimal recursion semantics \citep{copestake2005minimal} and head-driven phrase structure grammar \citep{pollard1994head}.  This formalism, illustrated in Figure~\ref{fig:dm-example} (top, blue) has the appealing property that a sentence's meaning is represented as a labeled, directed graph.  Vertices are words (though not every word is a vertex), and 59 labels are used to characterize argument and adjunct relationships, as well as conjunction.

Other semantic formalisms such as PSD \cite{hajic2012announcing}, EDS \citep{oepen2006discriminant}, and UCCA \cite{abend2013universal} also capture semantics as graphs.  Preliminary experiments showed similar findings %
using these.  Frame-based predicate-argument representations such as those found in PropBank \citep{palmer-etal-2005-proposition} and FrameNet \citep{framenet} are not typically cast as graphs (rather as ``semantic role labeling''), but see \citet{surdeanu-etal-2008-conll} for data transformations and \citet{peng2018learning} for methods that help bridge the gap.

Graph-based formalizations of predicate-argument semantics, along with organized shared tasks on semantic dependency parsing \citep{oepen2014semeval,oepen2015semeval}, enabled the development of data-driven parsing methods following extensive algorithm development for dependency \emph{syntax} \citep{eisner-1996-three,mcdonald-etal-2005-non}.  Even before the advent of the pretraining-finetuning paradigm, labeled $F_1$ scores above 0.9 were achieved \citep{peng2017deep}.

Some similarities between DM and dependency syntax (e.g., the Stanford dependencies, illustrated in Figure~\ref{fig:dm-example}, bottom, red; \citealp{de-marneffe-etal-2006-generating}) are apparent:  both highlight \emph{bilexical} relationships. However, semantically empty words (like infinitival \emph{to}) are excluded from the semantic graph, allowing direct connections between semantically related pairs
(e.g., \emph{technique} $\leftarrow$ \emph{apply}, \emph{impossible} $\rightarrow$ \emph{apply}, and \emph{apply} $\rightarrow$ \emph{crops}, all of which are mediated by other words in the syntactic graph). DM analyses need not be trees as in most syntactic dependency representations,\footnote{The enhanced universal dependencies of \citet{Schuster2016EnhancedEU} are a counterexample.} so they may more directly capture the meaning of many constructions, such as control.

\section{Probing RoBERTa for Predicate-Argument Semantics} \label{sec:probing}

The methodology known as ``linguistic probing'' seeks to determine the level to which a pretrained model has rediscovered a particular linguistic abstraction from raw data \interalia{shi-etal-2016-string, adi2016fine, ijcai2018-796, belinkov-glass-2019-analysis}.
The procedure is:
\begin{compactitem}
    \item[1.] Select an annotated dataset that encodes the theoretical abstraction of interest into a predictive task, usually mapping sentences to linguistic structures.  Here we will consider the Penn Treebank \citep{marcus-etal-1993-building} converted to Stanford dependencies and the DM corpus from SemEval 2015 Task 18 \citep{oepen2015semeval}.\footnote{These are both derived from the same \emph{Wall Street Journal} corpus and have similar size: the syntactic dependency dataset has 39,832/2,416 training/test examples, while the DM dataset has 33,964/1,410.}
    \item[2.] Pretrain.  We consider RoBERTa and BERT.
    \item[3.] \rev{Train a full-fledged ``ceiling'' model with finetuned representations. 
    It can be seen as proxy to the best performance one can get with the pretrained representations.}
    \item[4.] Train a supervised ``probe'' model for the task with the pretrained representations.
    Importantly, \rev{the pretrained representations should be \emph{frozen},
    and the probe model should be lightweight with limited capacity, 
    so that its performance is attributable to pretraining.
    We use a linear probe classifier.}
    \item[5.] \rev{
        Compare, on held-out data, the probe model against the ceiling model.
        Through such a comparison,
    }
    we can estimate the extent to which the pretrained model ``already knows'' how to do the task, or, more precisely, brings relevant features to the surface for use by the probing model.
\end{compactitem}

\citet{liu2019linguistic} included isolated DM arc prediction and labeling tasks and \citet{tenney2019you} conducted ``edge probing.'' %
To our knowledge, full-graph semantic dependency parsing has not been formulated as a probe.

\begin{table*}[t]
\begin{subtable}[tbh]{\textwidth}
\centering
\begin{tabular}{@{} l c@{\hspace{7pt}}c@{\hspace{7pt}}c@{\hspace{7pt}}c c@{\hspace{7pt}}c@{\hspace{7pt}}c@{\hspace{7pt}}c @{}}
\toprule

& \multicolumn{4}{c}{\textbf{PTB SD}} & \multicolumn{4}{c}{\textbf{SemEval 2015 DM}} \\
\cmidrule(lr){2-5} 
\cmidrule(lr){6-9}
\textbf{Metrics}
& \textbf{Abs} $\boldsymbol{\Delta}$ & \textbf{Rel} $\boldsymbol{\Delta}$ & \textbf{Ceiling} & \textbf{Probe}
& \textbf{Abs} $\boldsymbol{\Delta}$ & \textbf{Rel} $\boldsymbol{\Delta}$ & \textbf{Ceiling} & \textbf{Probe} \\
\midrule

\textbf{LAS/$F_1$} & --13.5\rev{$_{\pm 0.2}$} & --14.2\%\rev{$_{\pm 0.2}$} & 95.2\rev{$_{\pm 0.1}$} & 81.7\rev{$_{\pm 0.1}$} & 
--23.5\rev{$_{\pm 0.1}$} & --24.9\%\rev{$_{\pm 0.2}$} & 94.2\rev{$_{\pm 0.0}$} & 70.7\rev{$_{\pm 0.2}$} \\
\textbf{LEM} & --36.4\rev{$_{\pm 0.8}$} & --72.4\%\rev{$_{\pm 1.1}$} & 50.3\rev{$_{\pm 0.5}$} & 13.9\rev{$_{\pm 0.5}$} & 
--45.4\rev{$_{\pm 1.1}$} & --93.5\%\rev{$_{\pm 0.5}$} & 48.5\rev{$_{\pm 1.2}$} & \phantom{0}3.1\rev{$_{\pm 0.2}$} \\
\textbf{UEM} & --46.3\rev{$_{\pm 0.7}$} & --73.2\%\rev{$_{\pm 0.5}$} & 63.3\rev{$_{\pm 0.8}$} & 17.0\rev{$_{\pm 0.3}$} & 
--48.8\rev{$_{\pm 1.0}$} & --92.8\%\rev{$_{\pm 0.5}$} & 52.6\rev{$_{\pm 1.0}$} & \phantom{0}3.8\rev{$_{\pm 0.2}$} \\

\bottomrule

\end{tabular}
\caption{Base.}
\end{subtable}
\begin{subtable}[tbh]{\textwidth}
\centering
\color{\revcolor}
\begin{tabular}{@{} l c@{\hspace{7pt}}c@{\hspace{7pt}}c@{\hspace{7pt}}c c@{\hspace{7pt}}c@{\hspace{7pt}}c@{\hspace{7pt}}c @{}}
\toprule

& \multicolumn{4}{c}{\textbf{PTB SD}} & \multicolumn{4}{c}{\textbf{SemEval 2015 DM}} \\
\cmidrule(lr){2-5} 
\cmidrule(lr){6-9}
\textbf{Metrics}
& \textbf{Abs} $\boldsymbol{\Delta}$ & \textbf{Rel} $\boldsymbol{\Delta}$ & \textbf{Ceiling} & \textbf{Probe}
& \textbf{Abs} $\boldsymbol{\Delta}$ & \textbf{Rel} $\boldsymbol{\Delta}$ & \textbf{Ceiling} & \textbf{Probe} \\
\midrule

\textbf{LAS/$F_1$} & --17.6$_{\pm 0.1}$ & --18.5\%$_{\pm 0.1}$ & 95.3$_{\pm 0.0}$ & 77.7$_{\pm 0.1}$ &
--26.7$_{\pm 0.3}$ & --28.3\%$_{\pm 0.3}$ & 94.4$_{\pm 0.1}$ & 67.7$_{\pm 0.2}$ \\
\textbf{LEM} & --40.0$_{\pm 0.6}$ & --77.2\%$_{\pm 0.4}$ & 51.9$_{\pm 0.6}$ & 11.8$_{\pm 0.2}$ &
--46.6$_{\pm 1.1}$ & --94.4\%$_{\pm 0.1}$ & 49.3$_{\pm 1.1}$ & \phantom{0}2.7$_{\pm 0.0}$ \\
\textbf{UEM} & --50.2$_{\pm 0.6}$ & --77.4\%$_{\pm 0.2}$ & 64.8$_{\pm 0.7}$ & 14.6$_{\pm 0.2}$ &
--50.0$_{\pm 1.1}$ & --93.9\%$_{\pm 0.2}$ & 53.2$_{\pm 1.0}$ & \phantom{0}3.3$_{\pm 0.1}$ \\

\bottomrule

\end{tabular}
\caption{Large.}
\end{subtable}
\caption{\label{tab:probing} The \rev{RoBERTa-base (top) and RoBERTa-large (bottom)} parsing results for the full ceiling model and the probe on the \rev{PTB Stanford Dependencies (SD)} test set and the SemEval 2015 Task 18 in-domain test set. 
We also report their absolute and relative differences (probe -- full). 
The smaller the magnitude of the difference, the more relevant content the pretrained model already encodes. 
We report the canonical parsing metric (LAS for PTB dependency and labeled $F_1$ for DM) and labeled/unlabeled exact match scores (LEM/UEM). 
\rev{All numbers are mean $\pm$ standard deviation across three seeds.}
}
\end{table*}

For both syntactic and semantic parsing, our full ceiling model and our probing model are based on the \citet{dozat2016deep,dozat2018simpler} parser that underlies many state-of-the-art systems \interalia{clark-etal-2018-semi,li-etal-2019-sjtu}.  
Our ceiling model contains nonlinear multilayer perceptron (MLP) layers between RoBERTa/BERT and the arc/label classifiers, as in the original parser, and finetunes the pretrained representations.
The probing model, trained on the same data, freezes the representations and removes the MLP layers, 
yielding a linear model with limited capacity.
We measure the conventionally reported metrics: labeled attachment score for dependency parsing and labeled $F_1$ for semantic parsing, as well as labeled and unlabeled exact match scores. 
We \rev{follow the standard practice and use the Chu-Liu-Edmonds algorithm~\citep{chu-liu1965shortest,edmonds1967optimum} to decode the syntactic dependency trees and greedily decode the semantic graphs with local edge/label classification decisions}.
See Appendix~\ref{sec:hyperparameters} for training details. 

Comparisons between absolute scores on the two tasks are less meaningful.  Instead, we are interested in the \emph{difference} between the probe (largely determined by pretrained representations) and the ceiling (which benefits also from finetuning).
Prior work leads us to expect that the semantic probe will exhibit a larger difference than the syntactic one, signalling that pretraining surfaces syntactic abstractions more readily than semantic ones.
This is exactly what we see in  Tables~\ref{tab:probing} across all metrics, \rev{for both RoBERTa-base and RoBERTa-large, where all relative differences (probe -- full) are greater in magnitude for parsing semantics than syntax}. \rev{
Surprisingly, RoBERTa-large achieves worse semantic and syntactic probing performance than its base-sized counterpart across all metrics. 
This suggests that larger pretrained representations do not necessarily
come with better structural information for downstream models to exploit.
}
In Appendix~\ref{sec:bert-probing}, we also show that BERT-base shows the same qualitative pattern. 

\section{Finetuning with Semantic Graphs} \label{sec:method}

\begin{figure}[tbh]
	\centering
	\includegraphics[width=0.45\textwidth]{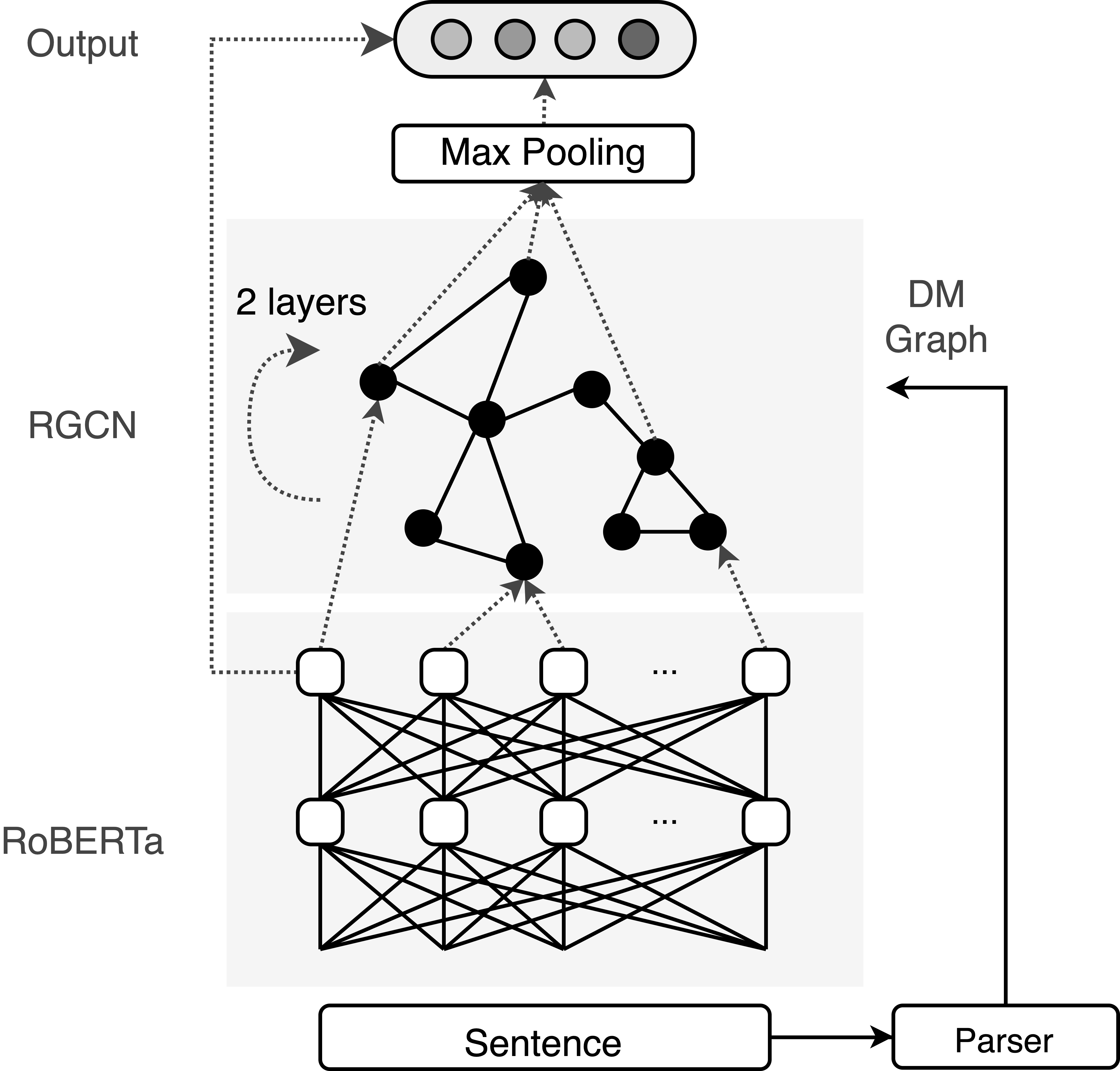}
	\caption{\model architecture.
	The sentence is first contextualized using RoBERTa, and then parsed.
	RGCN encodes the graph structures on top of RoBERTa.
	We max-pool over the RGCN's outputs for onward computation.
	}
	\label{fig:architecture}
\end{figure}

\begin{figure}[t]
	\centering
	\includegraphics[width=0.47\textwidth]{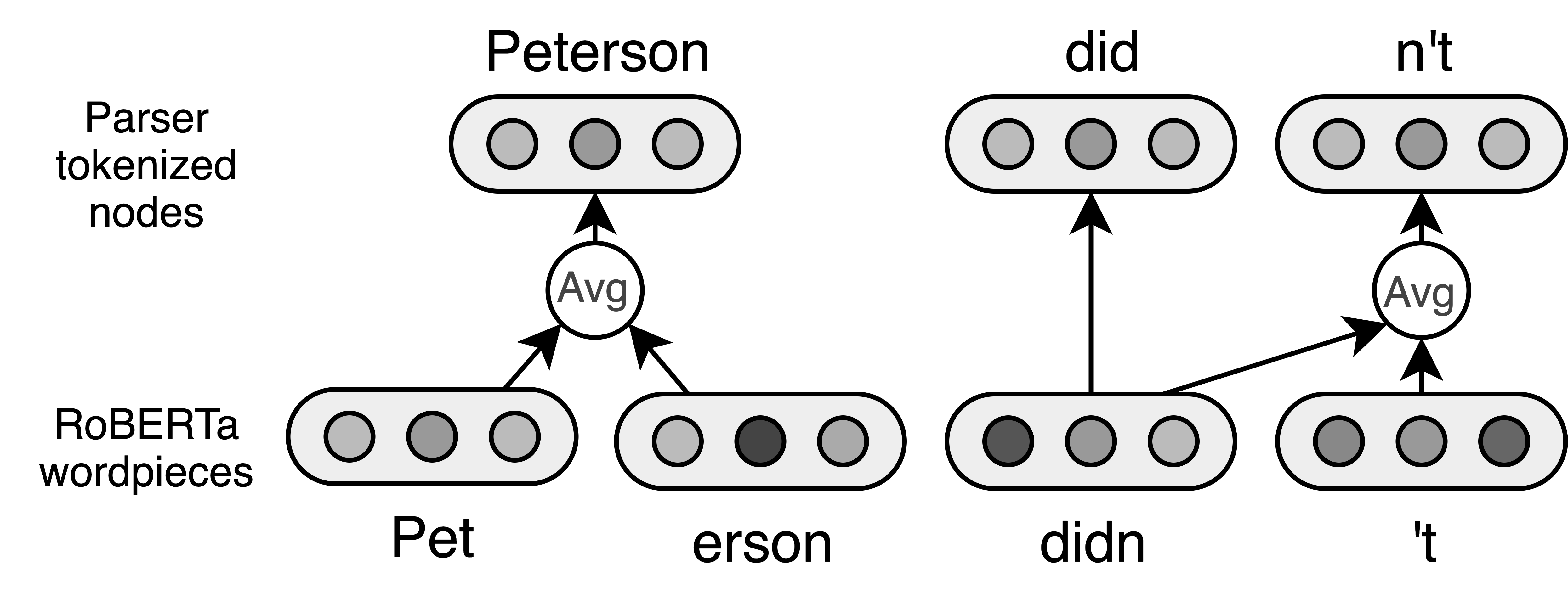}
	\caption{
	To get the representation of a node, we average the vectors of the wordpieces it is aligned to.
	}
	\label{fig:alignment}
\end{figure}

Given pretrained RoBERTa's relative incapability of surfacing semantic structures (\S\ref{sec:probing}) and the importance of modeling predicate-argument semantics (\S\ref{sec:pas}), we hypothesize that incorporating such information into the RoBERTa finetuning process should benefit downstream NLU tasks.

\model, briefly outlined in~\S\ref{sec:fuse}, is based on the relational graph convolutional network (RGCN; \citealp{schlichtkrull2018modeling}).
\S\ref{sec:fuse-light} introduces a lightweight variant of \model aiming to reduce test time memory and runtime.

\subsection{\model}\label{sec:fuse}
\model first uses an external parser to get the semantic analysis for the input sentence.
Then it contextualizes the input with a pretrained RoBERTa model,
the output of which is fed into a graph encoder building on the semantic parse.
We use RGCN to encode the DM structures, which are labeled graphs.
The model is trained end-to-end.
Figure~\ref{fig:architecture} diagrams this procedure.

\paragraph{RGCN.}
RGCN can be understood as passing vector ``messages'' among vertices in the graph.
The nodes are initially represented with RoBERTa token embeddings. 
At each RGCN layer, each node representation is updated
with a learned composition function, taking as input the vector representations of the node's neighbors as well itself. 
Each DM relation type is associated with a separately parameterized composition function.
For tasks such as text classification or regression, we max-pool over the final RGCN layer's output
to obtain a sequence-level representation for onward computation. %
Readers are referred to Appendix~\ref{sec:detailed-architecture} and \citet{schlichtkrull2018modeling} for further details.

\paragraph{Note on Tokenization.} 
RoBERTa uses byte-pair encodings (BPE;~\citealp{sennrich-etal-2016-neural}), differing from the CoNLL 2019 tokenizer \cite{oepen-etal-2019-mrp} used by the parser.
To get each token's initial representation for RGCN, we average RoBERTa's output vectors for the BPE wordpieces that the token is aligned to (illustrated in Figure~\ref{fig:alignment}).

\subsection{\modell}\label{sec:fuse-light}
Inspired by the scaffold model of \citet{swayamdipta2018syntactic}, we introduce \modell, 
a lightweight variant of \model that aims to reduce time and memory overhead at test time.
During inference it does \emph{not} rely on explicit semantic structures
and therefore has the same computational cost as the RoBERTa baseline.

\modell learns two classifiers (or regressors): (1) a main linear classifier on top of RoBERTa $f_{\text{RoBERTa}}$; 
(2) an auxiliary classifier $f_{\text{RGCN}}$ based on \model.
They are separately parameterized at the classifier level, but share the same underlying RoBERTa.
They are trained on the same downstream task and jointly update the RoBERTa model.
At test time, we only use $f_{\text{RoBERTa}}$.
The assumption behind \modell is similar to the scaffold framework of \citet{swayamdipta2018syntactic}:
by sharing the RoBERTa parameters between the two classifiers,
the contextualized representations steer towards downstream classification with semantic encoding.
One key difference is that
\modell learns with two different architectures for the same task,
instead of using the multitask learning framework of \citet{swayamdipta2018syntactic}.
In \S\ref{sec:alternative-architectures}, we find that \modell outperforms a scaffold.

\subsection{Discussion} \label{sec:model-discussion}
Previous works have used GCN \cite{kipf2016semi}, a similar architecture, to encode \emph{unlabeled} syntactic structures~\citep{zhang-etal-2020-syntax,zhang2020sg};
\citet{marcheggiani2017encoding} and \citet{bastings-etal-2017-graph} additionally learned label information with bias terms.
\rev{We use RGCN to explicitly encode \emph{labeled} semantic graphs by
learning a separate projection matrix for each semantic relation.
Our analysis shows that it outperforms GCN, as well as alternatives such as 
multitask learning with parameter-sharing (\S\ref{sec:alternative-architectures})}.
However, this comes with a cost.
In RGCN, the number of parameters linearly increases with the number of relation types.\footnote{
In experiments we upper-bound the number of the parameters 
by imposing a low-rank constraint on the parameter matrices by construction.
See Appendix~\ref{sec:detailed-architecture}.}
In our experiments, on top of the 125M RoBERTa-base parameters, this adds approximately 3--118M
parameters to the model, depending on the hyperparameter settings (see Appendix~\ref{sec:hyperparameters}). \rev{On top of RoBERTa-large, which itself has 355M parameters, this adds 6--121M additional parameters.}
The inference runtime of \model is
1.41--1.79$\times$ RoBERTa's \rev{with the base size and 1.30--1.53$\times$ with the large size}.

\rev{\model incorporates semantic information only during finetuning.
Recent evidence suggests that structural information can be learned 
with specially-designed pretraining procedures.
For example,  \citet{swayamdipta2019shallow} pretrain with syntactic chunking,
requiring the entire pretraining corpus to be parsed which is computationally prohibitive at the scale of RoBERTa's pretraining dataset.
With a distillation technique, 
\citet{kuncoro2020syntactic} bake syntactic supervision into the pretraining objective.
Despite better accuracy on tasks that benefit from syntax,
they show that the obtained syntactically-informed model  \emph{hurts} the performance on other tasks, which could restrict its general applicability.
Departing from these alternatives,
\model augments general-purpose pretraining with task-specific structural finetuning,
an attractively modular and flexible solution.
}

\section{Experiments} \label{sec:experiments}

We next present experiments with SIFT to test our hypothesis that pretrained models for natural language understanding tasks benefit from explicit predicate-argument semantics.

\subsection{Settings} \label{sec:settings}

\begin{table}[t!]
\centering
\begin{tabular}{@{} l  l  r  r @{}}
\toprule
\textbf{Data} & \textbf{Task} & \textbf{|Train|} & \textbf{|Dev.|} \\
\midrule
\textbf{CoLA} %
& Acceptability & 8.5K & 1K \\
\textbf{MRPC} %
& Paraphrase  & 2.7K & 409 \\
\textbf{QNLI} & Entailment & 105K & 5.5K \\
\textbf{RTE} & Entailment & 2.5K & 278 \\
\textbf{SST-2} %
& Sentiment & 67K & 873 \\
\textbf{STS-B} %
& Similarity & 5.8K & 1.5K \\
\textbf{QQP} %
& Paraphrase & 363K & 40K \\
\textbf{MNLI} %
& Entailment & 392K & 9.8K \\
\bottomrule
\end{tabular}
\caption{\label{tab:dataset-stats} GLUE datasets and statistics. %
CoLA: \citet{warstadt2019neural};
MRPC: \citet{dolan-brockett-2005-automatically};
SST-2: \citet{socher-etal-2013-recursive};
STS-B: \citet{cer-etal-2017-semeval};
QQP: \citet{qqp};
MNLI: \citet{williams2017broad};
QNLI is compiled by GLUE's authors using \citet{rajpurkar-etal-2016-squad}. RTE is the concatenation of \citet{10.1007/11736790_9,rte2,giampiccolo-etal-2007-third,Bentivogli09thefifth}.
}
\end{table}

We use the GLUE datasets, a suite of tests targeting natural language understanding detailed in Table~\ref{tab:dataset-stats} \citep{wang2018glue}.\footnote{Following \citet{devlin2018bert}, we do not report WNLI results because it is hard to outperform the majority class baseline using the standard classification finetuning routine.}
Most are classification datasets, while STS-B considers regression. 
Among the classifications datasets, MNLI has three classes while others have two;
CoLA and SST-2 classify single sentences while the rest classify sentence pairs.
We follow  \citet{dodge2020fine} and \citet{vu2020exploring} and only report development set results due to restricted GLUE test set access.

We compare the following models:
\begin{compactitem}
    \item\textbf{RoBERTa}, \rev{both the base and large variants}, following \citet{liu2019roberta}. %

    \item\textbf{\model} builds on pretrained RoBERTa, with 2 RGCN layers.  
    \rev{To generate semantic graphs, we use the semantic dependency parser by \citet{che2019hit} which held the first place in the CoNLL 2019 shared task \cite{oepen-etal-2019-mrp} with 92.5 labeled $F_1$ for DM.}\footnote{
    \rev{
    About half of the CoNLL 2019 evaluation set is out-of-domain.
    Without gold semantic graph annotations for our target datasets, this can be seen as a reasonable 
    estimation of the parser's performance for our use case.
    }}
    
    \item\textbf{\modell} (\S\ref{sec:fuse-light}) is trained similarly to \model, but does not rely on
    inference-time parsing. 
    
    \item\textbf{Syntax-infused finetuning} is similar to SIFT but uses the \emph{syntactic} Universal Dependencies parser \citep{straka-2018-udpipe,straka-strakova-2019-ufal} from the CoNLL 2019 shared task \cite{oepen-etal-2019-mrp}. 
    We include this model to confirm that any benefits to task performance are due specifically to the semantic structures.
\end{compactitem}
Hyperparameters are summarized in Appendix~\ref{sec:hyperparameters}.

\paragraph{Implementation Details.}
We run all models across 3 seeds for the large datasets QNLI, MNLI, and QQP (due to limited computational resources), 
and 4 seeds for all others.
As we do not aim for state of the art, we do \emph{not} use intermediate task training, ensemble models, or re-formulate QNLI as a ranking task as done by \citet{liu2019roberta}.
For sentence-pair classification tasks such as MNLI, we use structured decomposable 
attention \cite{parikh2016decomposable} and 2 additional RGCN layers to further propagate the attended information~\cite{chen2016enhanced}. The two graphs are separately max-pooled to obtain the final representation. See Appendix~\ref{sec:detailed-architecture} for more details.

\subsection{Main Findings} \label{sec:results}

\begin{table*}[t]
\begin{subtable}[tbh]{\textwidth}
\centering
\small
\begin{tabular}{@{} l @{\hspace{7pt}} c @{\hspace{8pt}} c @{\hspace{8pt}} c @{\hspace{8pt}} c @{\hspace{8pt}} c@{\hspace{8pt}} c@{\hspace{8pt}}c@{\hspace{8pt}} c@{\hspace{8pt}}c@{\hspace{7pt}} >{\color{\revcolor}}c @{}}
\toprule
&&&&&&&
&\multicolumn{2}{c}{\textbf{MNLI}} \\
\cmidrule(lr){9-10}

\textbf{Models}
&\textbf{CoLA}
&\textbf{MRPC}
&\textbf{RTE}
&\textbf{SST-2}
&\textbf{STS-B}
&\textbf{QNLI}
&\textbf{QQP}
& \textbf{ID.}
& \textbf{OOD.}
&\textbf{Avg.}\\

\midrule
\textbf{RoBERTa} & 63.1$_{\pm 0.9}$ & 90.1$_{\pm 0.8}$ & 79.0$_{\pm 1.6}$ 
& 94.6$_{\pm 0.3}$ & 91.0$_{\pm 0.0}$ & 93.0$_{\pm 0.3}$ & 91.8$_{\pm 0.1}$ & 87.7$_{\pm 0.2}$ & 87.3$_{\pm 0.3}$ & 86.4 \\

\midrule
\textbf{\model} & \textbf{64.8}$_{\pm 0.4}$ & 90.5$_{\pm 0.7}$ & 81.0$_{\pm 1.4}$ 
& 95.1$_{\pm 0.4}$ & \textbf{91.3}$_{\pm 0.1}$ & 93.2$_{\pm 0.2}$ & 91.9$_{\pm 0.1}$ & 87.9$_{\pm 0.2}$ & \textbf{87.7}$_{\pm 0.1}$ & 87.0 \\
\textbf{\modell} & 64.1$_{\pm 1.3}$ & 90.3$_{\pm 0.5}$ & 80.6$_{\pm 1.4}$ 
& 94.7$_{\pm 0.1}$ & \textbf{91.2}$_{\pm 0.1}$ & 92.8$_{\pm 0.3}$ & 91.7$_{\pm 0.0}$ & 87.7$_{\pm 0.1}$ & 87.6$_{\pm 0.1}$ & 86.7 \\

\midrule
\textbf{Syntax} & 63.5$_{\pm 0.6}$ & 90.4$_{\pm 0.5}$ & 80.9$_{\pm 1.0}$ 
& 94.7$_{\pm 0.5}$ & 91.1$_{\pm 0.2}$ & 92.8$_{\pm 0.2}$ & 91.8$_{\pm 0.0}$ & 87.9$_{\pm 0.1}$ & \textbf{87.7}$_{\pm 0.1}$ & 86.7 \\
\bottomrule

\end{tabular}
\caption{\label{tab:glue}Base.}
\end{subtable}
\begin{subtable}[tbh]{\textwidth}
\centering
\small
\color{\revcolor}
\begin{tabular}{@{} l @{\hspace{7pt}} c @{\hspace{8pt}} c @{\hspace{8pt}} c @{\hspace{8pt}} c @{\hspace{8pt}} c@{\hspace{8pt}} c@{\hspace{8pt}}c@{\hspace{8pt}} c@{\hspace{8pt}} c@{\hspace{7pt}} c @{}}
\toprule
&&&&&&&
&\multicolumn{2}{c}{\textbf{MNLI}} \\
\cmidrule(lr){9-10}

\textbf{Models}
&\textbf{CoLA}
&\textbf{MRPC}
&\textbf{RTE}
&\textbf{SST-2}
&\textbf{STS-B}
&\textbf{QNLI}
&\textbf{QQP}
& \textbf{ID.}
& \textbf{OOD.}
&\textbf{Avg.}\\

\midrule
\textbf{RoBERTa} & 68.0$_{\pm 0.6}$ & 90.1$_{\pm 0.8}$ & 85.1$_{\pm 1.0}$ 
& 96.1$_{\pm 0.3}$ & 92.3$_{\pm 0.2}$ & 94.5$_{\pm 0.2}$ & 91.9$_{\pm 0.1}$ & 90.3$_{\pm 0.1}$ & 89.8$_{\pm 0.3}$ & 88.7 \\

\midrule

\textbf{\model} & \textbf{69.7}$_{\pm 0.5}$ & \textbf{91.3}$_{\pm 0.4}$ & \textbf{87.0}$_{\pm 1.1}$ 
& 96.3$_{\pm 0.3}$ & \textbf{92.6}$_{\pm 0.0}$ & 94.7$_{\pm 0.1}$ & \textbf{92.1}$_{\pm 0.1}$ & 90.4$_{\pm 0.1}$ & 90.1$_{\pm 0.1}$ & 89.3 \\

\textbf{Syntax} & 69.6$_{\pm 1.2}$ & 91.0$_{\pm 0.5}$ & 86.0$_{\pm 1.6}$ 
& 95.9$_{\pm 0.3}$ & 92.4$_{\pm 0.1}$ & 94.6$_{\pm 0.1}$ & \textbf{92.0}$_{\pm 0.0}$ & 90.4$_{\pm 0.3}$ & 90.0$_{\pm 0.2}$ & 89.1 \\
\bottomrule
\end{tabular}
\caption{\label{tab:glue-large} Large.
}
\end{subtable}
\caption{\label{tab:glue-all} GLUE development set results \rev{with RoBERTa-base (top) and RoBERTa-large (bottom)}. 
We report Matthews correlation for CoLA, Pearson's correlation for STS-B, and accuracy for others. 
We report mean $\pm$ standard deviation;
\rev{for each bold entry, the mean minus standard deviation is no worse than RoBERTa's corresponding mean plus standard deviation.}
}
\end{table*}

\rev{Tables~\ref{tab:glue-all} summarizes the GLUE development set performance of the four aforementioned models when they are implemented with RoBERTa-base and RoBERTa-large}.
\rev{With RoBERTa-base (Table~\ref{tab:glue}),} 
\rev{\model  achieves a consistent improvement over the baseline across the board},
suggesting that despite heavy pretraining, RoBERTa still benefits from explicit semantic structural information. Among the datasets, smaller ones tend to obtain larger improvements from \model, e.g., 1.7 Matthews correlation for CoLA and 2.0 accuracy for RTE, 
while the gap is smaller on the larger ones (e.g., only 0.1 accuracy for QQP).
Moreover, \modell often improves over RoBERTa, with a smaller gap, making it a compelling model choice when latency is prioritized. 
This shows that encoding semantics using RGCN is not only capable of producing better standalone output representations, 
but can also benefit the finetuning of the RoBERTa-internal weights through parameter sharing.
Finally, the syntax-infused model underperforms \model across all tasks.
It only achieves minor improvements over RoBERTa, if not hurting performance. 
These results provide evidence supporting our hypothesis that incorporating semantic structures is more beneficial to RoBERTa than syntactic ones.

\rev{We observe a similar trend with RoBERTa-large in Table~\ref{tab:glue-large},
where \model's absolute improvements are very similar to those in Table~\ref{tab:glue}.
Specifically, both achieve an 0.6 accuracy improvement over RoBERTa, averaged across all datasets.
This indicates that the increase from RoBERTa-base to RoBERTa-large added little to surfacing semantic information.}

\section{Analysis and Discussion}\label{sec:analysis}

In this section, we first analyze in which scenarios incorporating semantic structures helps RoBERTa. We then highlight \model's data efficiency and compare it to alternative architectures. We show ablation results for architectural decisions in Appendix~\ref{sec:ablations}. \rev{All analyses are conducted on RoBERTa-base.}

\begin{table*}[tbh]
\centering
\small
\begin{tabular}{@{} l @{\hspace{6pt}} l @{\hspace{6pt}} l @{\hspace{0pt}} c @{\hspace{6pt}} c @{\hspace{6pt}} c @{}}
\toprule

\textbf{Heuristic} & \textbf{Premise} & \textbf{Hypothesis} & \textbf{Label} & \textbf{RoBERTa} & \textbf{\model} \\

\midrule
\multirow{2}{*}{\shortstack[l]{Lexical\\ Overlap}}
 & The banker near the judge saw the actor. & The banker saw the actor. & E & 98.3 & \textbf{98.9} \\
 & The judge by the actor stopped the banker. & The banker stopped the actor. & N & 68.1 & \textbf{71.0} \\
\midrule

\multirow{2}{*}{\shortstack[l]{Sub-\\ sequence}}
 & The artist and the student called the judge. & The student called the judge. & E & 99.7 & \textbf{99.8} \\
 & The judges heard the actors resigned. & The judges heard the actors. & N & 25.8 & \textbf{29.5} \\
\midrule
\multirow{2}{*}{Constituent}
 & Before the actor slept, the senator ran. & The actor slept. & E & \textbf{99.3} & 98.8 \\
& If the actor slept, the judge saw the artist. & The actor slept. & N & \textbf{37.9} & 37.6 \\
\bottomrule

\end{tabular}
\caption{\label{tab:hans-results} HANS heuristics and RoBERTa-base and \model's accuracy. 
Examples are due to \citet{mccoy2019right}.
``E'': entailment. ``N'': non-entailment. Bold font indicates the better result in each category.
}
\end{table*}

\begin{table}[tb]
\centering
\begin{tabular}{@{} l @{\hspace{8pt}} c @{\hspace{8pt}} c}
\toprule
\textbf{Phenomenon} & \textbf{RoBERTa} & \textbf{\model} \\ \midrule

Predicate Argument Structure & 43.5 & \textbf{44.6} \\
Logic & 36.2 & \textbf{38.3} \\
Lexical Semantics & \textbf{45.6} & 44.8 \\
Knowledge & \textbf{28.0} & 26.3 \\

\bottomrule

\end{tabular}
\caption{\label{tab:diagnostic-results} $R_3$ correlation coefficient of RoBERTa-base and \model on the GLUE diagnostic set.
}
\end{table}

\subsection{When Do Semantic Structures Help?}\label{sec:hans} 
Using two diagnostic datasets designed for \rev{evaluating and analyzing} natural language inference models,
we find that \model 
(1) helps guard the model against frequent but \emph{invalid} heuristics in the data, and
(2) better captures nuanced sentence-level linguistic phenomena than RoBERTa.

\paragraph{Results on the HANS Diagnostic Data.}

We first diagnose the model using the HANS dataset~\citep{mccoy2019right}. 
It aims to study whether a natural language inference (NLI) system adopts three heuristics, summarized  and exemplified in Table~\ref{tab:hans-results}.
The premise and the hypothesis have high surface form overlap, but the heuristics are \emph{not} valid for reasoning.
Each heuristic has both positive and negative (i.e., entailment and non-entailment) instances constructed. 
Due to the high surface similarity, many models tend to predict ``entailment'' for the vast majority of instances.
As a result, they often reach decent accuracy on the entailment examples,
but struggle on the ``non-entailment'' ones~\citep{mccoy2019right}, on which we focus our analysis.
The 30,000 test examples are evenly spread among the 6 classes (3 heuristics, 2 labels).

Table~\ref{tab:hans-results} compares \model against the RoBERTa baseline on HANS.  
Both struggle with non-entailment examples.
\model yields improvements on the lexical overlap and subsequence heuristics, which we find unsurprising, given that semantic analysis directly addresses the underlying differences in meaning between the (surface-similar) premise and hypothesis in these cases.
\model \rev{performs similarly to} RoBERTa on the constituent heuristic \rev{with a 0.3\% accuracy difference for the non-entailment examples}.  Here the hypothesis corresponds to a constituent in the premise, and therefore we expect its semantic parse to often be a subgraph of the premise's; accuracy hinges on the meanings of the connectives (e.g., \emph{before} and \emph{if} in the examples), not on the structure of the graphs. 

\paragraph{Results on the GLUE Diagnostic Data.}
GLUE's diagnostic set \citep{wang2018glue} contains 1,104 artificially curated NLI examples to test a model's performance on various linguistic phenomena including
\textbf{predicate-argument structure} (e.g., ``I opened the door.'' entails ``The door opened.'' but not ``I opened.''), 
\textbf{logic} (e.g., ``I have no pet puppy.'' entails ``I have no corgi pet puppy.'' but not ``I have no pets.''), 
\textbf{lexical semantics} (e.g., ``I have a dog.'' entails ``I have an animal.'' but not ``I have a cat.''), 
and \textbf{knowledge \& common sense} (e.g., ``I went to the Grand Canyon.'' entails ``I went to the U.S..'' but not ``I went to Antarctica.''). 
Table~\ref{tab:diagnostic-results} presents the results in $R_3$ correlation coefficient \cite{10.1016/j.compbiolchem.2004.09.006}.
Explicit semantic dependencies help \model perform better on predicate-argument structure and sentence logic.
On the other hand, \model underperforms the baseline on lexical semantics and world knowledge.  We would not expect a benefit here, since semantic graphs do not add lexical semantics or world knowledge; the drop in performance suggests that some of what RoBERTa learns is lost when it is finetuned through sparse graphs.  Future work might seek graph encoding architectures that mitigate this loss.

\subsection{Sample Efficiency} \label{sec:data-efficiency}
In \S\ref{sec:results}, we observe greater improvements from \model on smaller finetuning sets.
We hypothesize that the structured inductive bias helps \model more when the amount of finetuning data is limited.
We test this hypothesis on MNLI by training different models varying the amount of finetuning data.
We train all configurations with the same three random seeds. As seen in Table~\ref{tab:data-efficiency}, \model offers larger improvements when less finetuning data is used.
Given the success of the pretraining paradigm, we expect many new tasks to emerge with tiny finetuning sets, and these will benefit the most from methods like \model.

\begin{table*}[t]
\centering
\begin{tabular}{@{} r r  >{\color{\revcolor}}c>{\color{\revcolor}}ccc >{\color{\revcolor}}c>{\color{\revcolor}}ccc @{}}
\toprule

&& \multicolumn{4}{c}{\textbf{ID.}} & \multicolumn{4}{c}{\textbf{OOD.}} \\
\cmidrule(lr){3-6}
\cmidrule(lr){7-10}

\textbf{Fraction} & \textbf{|Train|} & \textbf{RoBERTa} & \textbf{SIFT} & \textbf{Abs} $\boldsymbol{\Delta}$ & \textbf{Rel} $\boldsymbol{\Delta}$ & \textbf{RoBERTa} & \textbf{SIFT} & \textbf{Abs} $\boldsymbol{\Delta}$ & \textbf{Rel} $\boldsymbol{\Delta}$ \\
\midrule
100\% & 392k & 87.7 & 87.9 & 0.2 & 0.2\% & 87.3 & 87.7 & 0.4 & 0.4\% \\
0.5\% & 1,963 & 76.1 & 77.6 & 1.5 & 1.9\% & 77.1 & 78.2 & 1.1 & 1.4\% \\
0.2\% & 785 & 68.6 & 71.0 & 2.5 & 3.5\% & 70.0 & 71.8 & 1.8 & 2.5\% \\
0.1\% & 392 & 58.7 & 61.2 & 2.6 & 4.2\% & 60.5 & 63.7 & 3.3 & 5.1\% \\
\bottomrule
\end{tabular}
\caption{\label{tab:data-efficiency} 
\rev{RoBERTa-base and \model's performance on the entire MNLI development sets
and their absolute and relative differences,}
with different numbers of finetuning instances randomly subsampled from the training data. }
\end{table*}

\subsection{Comparisons to Other Graph Encoders} \label{sec:alternative-architectures}

\begin{table*}[tbh]
\centering
\begin{tabular}{@{} l @{\hspace{7pt}} ccccccc cc c @{}}
\toprule
&&&&&&&
&\multicolumn{2}{c}{\textbf{MNLI}} \\
\cmidrule(lr){9-10}

\textbf{Models}
&\textbf{CoLA}
&\textbf{MRPC}
&\textbf{RTE}
&\textbf{SST-2}
&\textbf{STS-B}
&\textbf{QNLI}
&\textbf{QQP}
& \textbf{ID.}
& \textbf{OOD.}
& \textbf{Avg.}\\

\midrule
\textbf{RoBERTa} & 63.1 & 90.1 & 79.0 
& 94.6 & 91.0 & 93.0 & 91.8 & 87.7 & 87.3 & 86.4 \\

\midrule
\textbf{GCN} & \textbf{65.2} & 90.2 &80.2 & 94.8 & 91.1 & 92.9 & 91.8 & 87.8 & \textbf{87.7} & 86.8 \\
\textbf{GAT} & 63.4 & 90.0 & 79.4 & 94.7 & 91.2 & 92.9 & 91.8 & 87.7 & 87.6 & 86.5 \\
\textbf{Hidden} & 64.2 & 90.2 & 79.7 & 94.5 & 91.0 & 92.8 & 91.8 & 87.1 & 86.7 & 86.4 \\
\textbf{Scaffold} & 62.5 & \textbf{90.5} & 71.1 & 94.3 & 91.0 & 92.6 & 91.7 & 87.7 & 87.6 & 85.5 \\
\midrule
\textbf{\model} & 64.8 & \textbf{90.5} & \textbf{81.0} 
& \textbf{95.1} & \textbf{91.3} & \textbf{93.2} & \textbf{91.9} & \textbf{87.9} & \textbf{87.7} & \textbf{87.0} \\
\textbf{\modell} & 64.1 & 90.3 & 80.6 
& 94.7 & 91.2 & 92.8 & 91.7 & 87.7 & 87.6 & 86.7 \\

\bottomrule

\end{tabular}
\caption{\label{tab:alternative-architectures} GLUE development set results for different architectures for incorporating semantic information. The settings and metrics are identical to Table~\ref{tab:glue}. All models use the base size variant.
}
\end{table*}

In this section we compare RGCN to some commonly used graph encoders.
We aim to study whether or not
(1) encoding graph labels helps,
and (2) explicitly modeling discrete structures is necessary.
Using the same experiment setting as in \S\ref{sec:settings},
we compare \model and \modell to
\begin{compactitem}
    \item Graph convolutional network~(\textbf{GCN};~\citealp{kipf2016semi}). 
    GCN does \emph{not} encode relations, but is otherwise the same as RGCN.
    \item Graph attention network~(\textbf{GAT};~\citealp{velivckovic2017graph}). 
    Similarly to GCN, it encodes \emph{unlabeled} graphs.
    Each node aggregates representations of its neighbors using an attention function (instead of convolutions). 
    \item \textbf{Hidden} \citep{pang2019improving, zhang-etal-2020-syntax}.
    It does \emph{not} explicitly encode structures, but uses the hidden representations
    from a pretrained parser as additional features to the classifier. 
    \item \textbf{Scaffold} \citep{swayamdipta2018syntactic} is based on
    multitask learning. It aims to improve the downstream task
    performance by additionally training the model on the DM data with a full parsing objective.
\end{compactitem}
To ensure fair comparisons, we use comparable implementations for these models.
We refer the readers to the works cited for further details. 

Table~\ref{tab:alternative-architectures} summarizes the results, with \model having the highest average score across all datasets. Notably, the 0.2 average absolute benefit of SIFT over GCN and 0.5 over GAT demonstrates the benefit of including the semantic relation types (labels).  Interestingly, on the linguistic acceptability task---which focuses on well-formedness and therefore we expect relies more on syntax---GCN outperforms RGCN-based SIFT.
GAT underperforms GCN by 0.3 on average, likely because the sparse semantic structures (i.e., small degrees of each node) make attended message passing less useful.  Hidden does not on average outperform the baseline, highlighting the benefit of discrete graph structures (which it lacks).  Finally, the scaffold underperforms across most tasks.

\section{Related Work}

\paragraph{Using Explicit Linguistic Information.} 
Before pretrained contextualized representations emerged, linguistic information was commonly incorporated into deep learning models to improve their performance including part of speech \interalia{sennrich2016linguistic, xu2016question} and syntax \interalia{eriguchi2017learning, chen2016enhanced, miwa2016end}. 
Nevertheless, recent attempts in incorporating syntax into pretrained models have had little success on NLU: 
\citet{strubell2018linguistically} found syntax to only marginally help semantic role labeling with ELMo, and \citet{kuncoro2020syntactic} observed that incorporating syntax into BERT conversely hurts the performance on some GLUE NLU tasks.
On the other hand, fewer attempts have been devoted to incorporating sentential predicate-argument semantics into NLP models. 
\citet{zhang2019semantics} embedded semantic role labels from a pretrained parser to improve BERT. 
However, these features do not constitute full sentential semantics. 
\citet{peng2018backpropagating} enhanced a sentiment classification model with DM but only used one-hop information and no relation modeling.

\paragraph{Probing Syntax and Semantics in Models.} 
Many prior works have probed the syntactic and semantic content of pretrained transformers, typically BERT. 
\citet{wallace-etal-2019-nlp} observed that BERT displays suboptimal numeracy knowledge. 
\citet{clark-etal-2019-bert} discovered that BERT's attention heads tend to surface syntactic relationships. 
\citet{hewitt2019structural} and \citet{tenney2019you} both observed that BERT embeds a significant amount of syntactic knowledge. \rev{Besides pretrained transformers, \citet{belinkov2020cl} used syntactic and semantic dependency relations to analyze machine translation models.}

\section{Conclusion}

We presented strong evidence that RoBERTa and BERT do not bring predicate-argument semantics to the surface as effectively as they do for syntactic dependencies.
This observation motivates \model, which aims to incorporate explicit semantic structures into the pretraining-finetuning paradigm.
It encodes automatically parsed semantic graphs using RGCN.  In controlled experiments, we find consistent benefits across eight 
tasks targeting natural language understanding, relative to RoBERTa and a syntax-infused RoBERTa. 
These findings motivate continued work on task-independent semantic analysis, including training methods that integrate it into architectures serving downstream applications.

\section*{Acknowledgments}
The authors thank the anonymous reviewers for feedback that improved the paper. We also thank Stephan Oepen %
for help in producing the CoNLL 2019 shared task companion data,
Yutong Li for contributing to early experiments,
and Elizabeth Clark and Lucy Lin for their suggestions and feedback. 
This research was supported in part by a Google Fellowship to HP and NSF grant 1562364.

\bibliography{main}
\bibliographystyle{acl_natbib}

\clearpage
\appendix

\section{Detailed Model Architecture} \label{sec:detailed-architecture}

In this section we provide a detailed illustration of our architecture.

\paragraph{Graph Initialization} 
Because RoBERTa's BPE tokenization differs from the \citet{che2019hit} semantic parser's CoNLL 2019 tokenization, we align the two tokenization schemes using character level offsets, as illustrated in Figure~\ref{fig:alignment}.
For each node $i$, we find wordpieces $[t_j, \cdots, t_k]$ that it aligns to.
We initialize its node embedding by averaging the vectors of these wordpiece followed by an
learned affine transformation and a $\operatorname{ReLU}$ nonlinearity:
\begin{align*}
    \vh^{(0)}_i = \operatorname{ReLU}\left(\mW_e \frac{1}{k - j + 1} \sum_{s=j}^k \ve_s\right)
\end{align*}
Here $\mW_e$ is a learned matrix, and the $\ve$ vectors are the wordpiece representations. 
The superscript on $\vh$ denotes the layer number, with 0 being the input embedding vector fed into the RGCN layers.

\paragraph{Graph Update} 
In each RGCN layer $\ell$, every node's hidden representation is propagated to its direct neighbors:
\begin{align*}
    &\vh_i^{(\ell + 1)} =\\
    &\operatorname{ReLU}\left(\sum_{r \in \gR} \sum_{j \in \gN_i^r} \frac{1}{\abs{\gN_i^r}} \mW_r^{(\ell)} \vh_j^{(\ell)} + \mW_0^{(\ell)} \vh_i^{(\ell)}\right)
\end{align*}
where $\gR$ is the set of all possible relations (i.e., edge labels; including inverse relations for inverse edges that we manually add corresponding to the original edges) and $\gN_i^r$ denotes $v_i$'s neighbors with relation $r$. 
$\mW_r$ and $\mW_0$ are learned parameters representing a relation-specific transformation and a self-loop transformation, respectively. We also use the basis-decomposition trick described in \citet{schlichtkrull2018modeling} to reduce the number of parameters and hence the memory requirement. 
Specifically, we construct $B$ basis matrices;
where $\abs{\gR} > B$, the transformation of each relation is constructed by a learned linear combination of the basis matrices. 
Each RGCN layer captures the neighbors information that is one hop away. 
We use $\ell = 2$ RGCN layers for our experiments.

\begin{figure}
	\centering
	\includegraphics[width=0.48\textwidth]{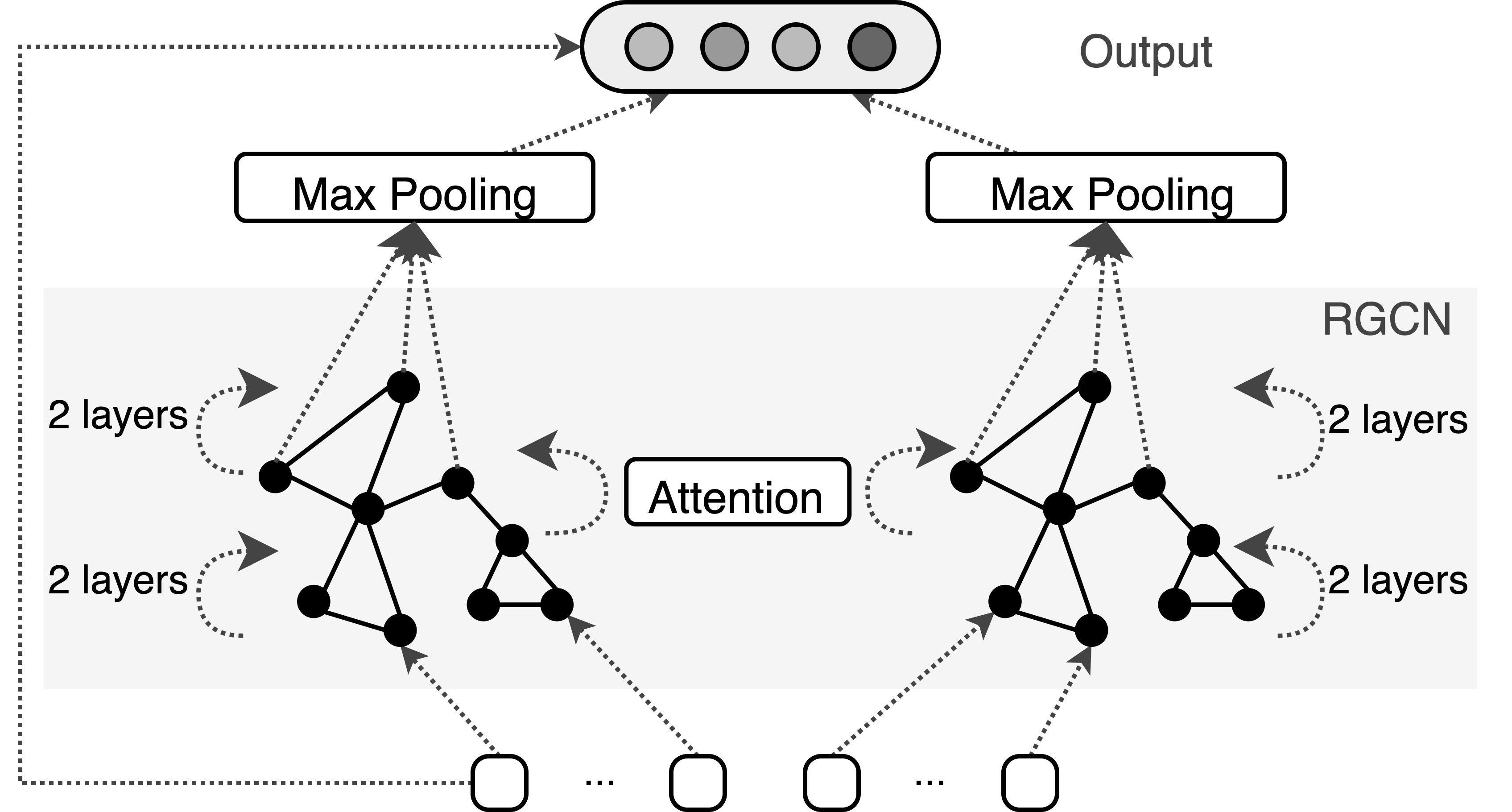}
	\caption{\model architecture for sentence pair tasks. 
	Two graphs are first separately encoded using RGCN, then  structured decomposable attention is used to capture the inter-graph interaction.
	Additional RGCN layers are used to further propagate the structured information.
	Finally two vectors max-pooled from both graphs are concatenated and used for onward computation.
	RoBERTa and the external parser are suppressed for clarity.
	}
	\label{fig:sentence_pair_se}
\end{figure}

\begin{table*}[tbh]
\centering
\begin{tabular}{@{} l @{\hspace{10pt}} c@{\hspace{7pt}}c@{\hspace{7pt}}c@{\hspace{7pt}}c  c@{\hspace{7pt}}c@{\hspace{7pt}}c@{\hspace{7pt}}c @{}}
\toprule

& \multicolumn{4}{c}{\textbf{PTB SD}} & \multicolumn{4}{c}{\textbf{SemEval 2015 DM}} \\
\cmidrule(lr){2-5} 
\cmidrule(lr){6-9}
\textbf{Metrics} & \textbf{Abs} $\boldsymbol{\Delta}$ & \textbf{Rel} $\boldsymbol{\Delta}$ & \textbf{Full} & \textbf{Probe} & \textbf{Abs} $\boldsymbol{\Delta}$ & \textbf{Rel} $\boldsymbol{\Delta}$ & \textbf{Full} & \textbf{Probe} \\
\midrule

\textbf{LAS/$F_1$} & --13.6 & --14.4\% & 94.6 & 81.0
& --23.2 & --24.8\% & 93.6 & 70.4 \\
\textbf{LEM} & --35.8 & --73.7\% & 48.6 & 12.8
& --39.4 & --91.6\% & 43.0 & \phantom{0}3.6 \\
\textbf{UEM} & --44.7 & --74.1\% & 60.3 & 15.7
& --42.0 & --91.5\% & 45.9 & \phantom{0}3.9 \\

\bottomrule

\end{tabular}
\caption{\label{tab:bert-probing} The BERT-base parsing results for the full ceiling model and the probing model on the PTB Stanford Dependencies (SD) test set and the SemEval 2015 Task 18 in-domain test set. The metrics and settings are identical to Table~\ref{tab:probing} except only one seed is used.}
\end{table*}

\begin{table}[t]
\centering
\begin{tabular}{@{}l c c c c @{}}
\toprule
&&& \multicolumn{2}{c}{\textbf{MNLI}} \\
\cmidrule(lr){4-5}
& \textbf{MRPC} & \textbf{STS-B} & \textbf{ID.} & \textbf{OOD.} \\ 

\midrule

\textbf{Full} & \textbf{90.5} & \textbf{91.3} & \textbf{87.9} & \textbf{87.7} \\
\textbf{-- attention} & 90.1 & 91.2 & \textbf{87.9} & \textbf{87.7} \\
\textbf{-- concat} & 90.2 & 91.0 & 87.8 & 87.6 \\

\bottomrule

\end{tabular}
\caption{\label{tab:ablations} Ablation results on the development sets of 3 GLUE datasets with a RoBERTa-base backbone.}
\end{table}

\paragraph{Sentence Pair Tasks}

For sentence pair tasks, it is crucial to model sentence interaction \citep{parikh2016decomposable}. 
We therefore use a similar structured decomposable attention component to model the interaction between the two semantic graphs. 
Each node attends to the other graph's nodes using biaffine attention;
its output is then concatenated to its node representation calculated in its own graph.
Specifically,
for two sentences $a$ and $b$, we obtain an updated representation $\vh'^{(\ell), a}$ for $a$ as follows:
\begin{align*}
    \alpha_{i, j} &= \operatorname{biaffine}\big(\vh_i^{(\ell), a}, \vh_j^{(\ell), b}\big)\\
    \tilde{\vh}_i^{(\ell), a} &= \sum_j \operatorname{softmax}_j(\alpha_{i, j}) \vh_j^{(\ell), b}\\
    \vh_i'^{(\ell), a} &= \operatorname{ReLU}\Big(\mW_\alpha\\
    \span [\vh_i^{(\ell), a}; \tilde{\vh}_i^{(\ell), a}; \vh_i^{(\ell), a} - \tilde{\vh}_i^{(\ell), a}; \vh_i^{(\ell), a} \odot \tilde{\vh}_i^{(\ell), a}]\Big)
\end{align*}
where $\mW_\alpha$ is a learned matrix,
and $\odot$ denotes the elementwise product.
We do the same operation to obtain the updated $\vh_j'^{(\ell), b}$.
Inspired by \citet{chen2016enhanced}, we add another $\ell$ RGCN composition layers to further propagate the attended representation. They result in additional parameters and runtime cost compared to what was presented in \S\ref{sec:model-discussion}.

\paragraph{Graph Pooling}

The NLU \rev{tasks} we experiment with require one vector representation for each instance.
We max-pool over the sentence graph (for sentence pair tasks, separately for the two graphs whose pooled output are then concatenated), concatenate it with RoBERTa's \emph{[CLS]} embedding, and feed the result into a layer normalization layer to get the final output.

\section{Hyperparameters} \label{sec:hyperparameters}

\paragraph{Probing Hyperparameters.} 
No hyperparameter tuning is conducted for the probing experiments. For the full models, we use intermediate MLP layers with dimension 512 for arc projection and 128 for label projection. The probing models do not have such layers. We minimize the sum of the arc and label cross entropy losses for both dependency and DM parsing. All models are optimized with AdamW~\citep{loshchilov2018decoupled} for 10 epochs with batch size 8 and learning rate $2 \times 10^{-5}$.

\paragraph{Main Experiment Hyperparameters.} For \model, we use 2 RGCN layers for single-sentence tasks and 2 additional composition RGCN layers after the structured decomposable attention component for sentence-pair tasks. The RGCN hidden dimension is searched in $\{256, 512, 768\}$, the number of bases in $\{20, 60, 80, 100\}$, dropout between RGCN layers in $\{0, 0.2, 0.3\}$, and the final dropout after all RGCN layers in $\{0, 0.1\}$. For \modell, the training loss is obtained with $0.2 \text{loss}_\text{RGCN} + 0.8 \text{loss}_\text{RoBERTa}$. For all models, the number of training epochs is searched in $\{3, 10, 20\}$ and the learning rate in $\{1 \times 10^{-4}, 2 \times 10^{-5}\}$. We use 0.1 weight decay and 0.06 warmup ratio. All models are optimized with AdamW with an effective batch size of 32.

\section{BERT Probing Results} \label{sec:bert-probing}

We replicate the RoBERTa probing experiments described in \S\ref{sec:probing} for BERT. We observe similar trends where the probing model degrades more from the full model for DM than dependency syntax. This demonstrates that, like RoBERTa, BERT also less readily surfaces semantic content than syntax.

\section{Ablations} \label{sec:ablations}

In this section we ablate two major architectural choices: the sentence pair structured decomposable attention component and the use of a concatenated RoBERTa and RGCN representation rather than only using the latter. We select 3 sentence-pair datasets covering different dataset sizes and tasks with identical experimental setup as \S\ref{sec:settings}. The ablation results in Table~\ref{tab:ablations} show that the full \model architecture performs the best.

\end{document}